\documentclass[letterpaper, 10 pt, conference]{ms}  
\IEEEoverridecommandlockouts

\usepackage[T1]{fontenc}
\usepackage[cmex10]{amsmath}
\usepackage{amsfonts}
\usepackage{siunitx}
\usepackage{array}
\usepackage{graphicx}
\usepackage{subfigure,stfloats,array}
\usepackage{hhline}
\usepackage{multirow}
\usepackage{algorithm}
\usepackage{algpseudocode}
\usepackage{cite}
\usepackage{hyperref}
\usepackage{color}

\begin{document}
\title{Semi-supervised Image Classification \\ with Grad-CAM Consistency}

\author{Juyong~Lee$^{1*}$,~
        Seunghyuk~Cho$^{1}$%
\thanks{This work was based on project during class "Deep Learning" in Pohang University of Science and Technology (POSTECH).}
\thanks{$^{1}$J.Y Lee, S.H Cho are the members of the Department of Computer Science and Engineering, POSTECH, Pohang 37673, South Korea. {\tt\small gimme1dollar@postech.ac.kr}}%
}
\maketitle

\begin{abstract}
Consistency training, which exploits both supervised and unsupervised learning with different augmentations on image, is an effective method of utilizing unlabeled data in semi-supervised learning (SSL) manner. Here, we present another version of the method with Grad-CAM consistency loss, so it can be utilized in training model with better generalization and adjustability. We show that our method improved the baseline ResNet model with at most 1.44 \% and 0.31 $\pm$ 0.59 \%p accuracy improvement on average with CIFAR-10 dataset. We conducted ablation study comparing to using only psuedo-label for consistency training. Also, we argue that our method can adjust in different environments when targeted to different units in the model. The code is available: https://github.com/gimme1dollar/gradcam-consistency-semi-sup.
\end{abstract}

\IEEEpeerreviewmaketitle

\section{Introduction}
Deep learning algorithm has become the state-of-the-art in many computer vision tasks, yet demanding a large amount of labeled data \cite{dl}. So,  semi-supervised learning (SSL) \cite{SSL} has appeared to be an appealing method of utilizing unlabeled data, and one popular class of SSL method is consistency training \cite{consistency}. With this method, model obtains a predicted label from an input data, and uses it as pseudo-label on the input perturbed or model modified. Here, we present a method of consistency training with Grad-CAM loss, which has better generalization on prediction and adjustability for differnet datasets with different properties. 

The training scheme with saliency-map consistency constraint has been shown in \cite{ECT}, while it is not fully examined as it uses old explanation method. While \cite{cam_ct} uses developed version of class activation map (CAM), this also follows the same issue of old version. In our method, by using Grad-CAM \cite{gradcam}, we did not only train the model to produce robust results on perturbation, but also control the training procedure by attaching the Grad-CAM module in different layers. Since Grad-CAM module can be attached in any target layers in the architecture unlike other methods, it can be used to guide the model in different scales. 

We conducted several experiments on the baseline model ResNet \cite{resnet} with and without Grad-CAM consistency loss, and have shown improvement on image classification task. The ablation study is followed to show Grad-CAM consistency loss without any conventional consistency training is useful. Also, examination on the method by targeting different layers suggests better use-case of our method. 

\begin{figure}[!tb]
 	\centering
    	\includegraphics[width=7cm]{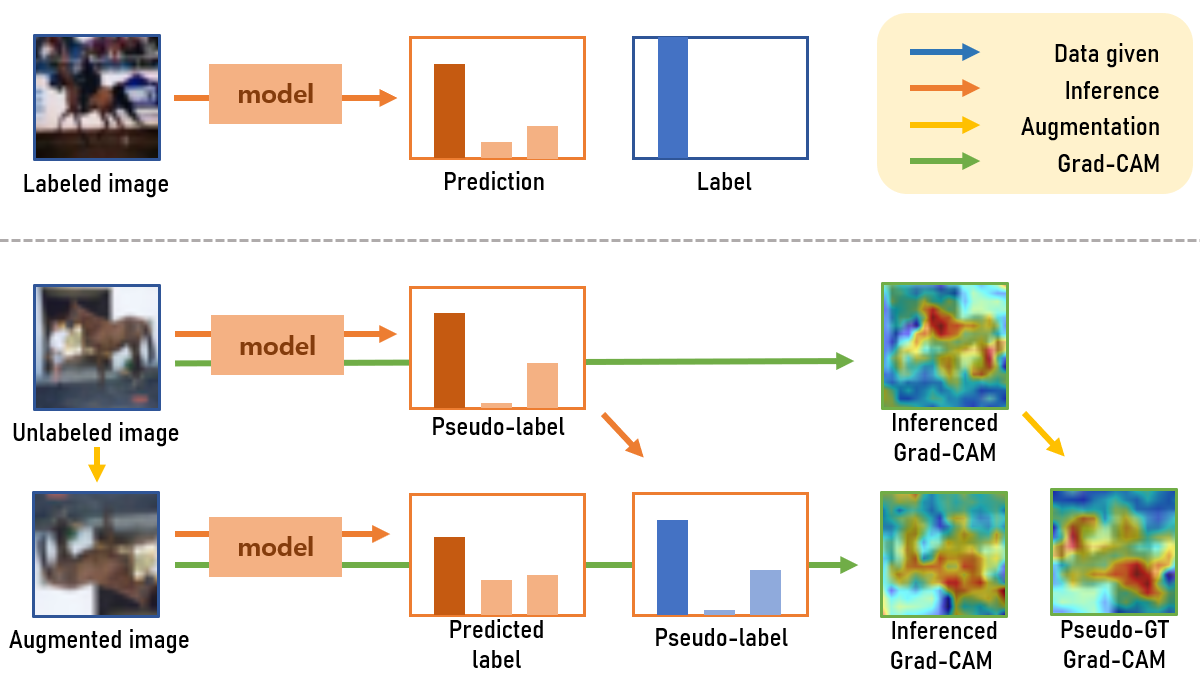}
    	\caption{ An overview of consistency training with Grad-CAM loss. While conventional supervised learning algorithms use label for supervision, consistency training uses pseudo-label predicted from original unlabeled image, and train the model to produce consistent result among the original and augmented image. In our setting, the model outputs consistent Grad-CAM results. }
      	\label{fig:overview}
 \end{figure}
 
\section{Related Works}

\textbf{Semi-supervised learning} is the main paradigm of unlabeled data exploitation. Many SSL applications are based on 'smoothness assumption' \cite{SSL_smooth}, meaning that similar data would have similar labels. Another example is entropy minimization strategy \cite{SSL_entropy}, leveraging 'low-density assumption' implying that the decision boundary of trained model does not pass high data-density region in the space. Similarly, 'manifold assumption' is also often used for SSL training \cite{SSL_manifold}, so that the the data is assumed to be lied on a low dimensional manifold. 

\textbf{Consistency training} is one of the most popular SSL approaches~\cite{MixMatch, meanteacher, VAT}. During the training, the model is trained to produce consistent results on inputs with various perturbations (e.g, crop or jittering). This method is under the assumption of the class probability should be same, so minimizing the error between $ || p(y; x) - p(y; Aug(x)) ||_2 ^2  $ where $ x $ is the input, $ y $ is the output, and $ Aug( \cdot ) $ is a stochastic transformation function with invariant label. There can be variety on using the loss function, as cross-entropy loss is chosen in FixMatch \cite{fixmatch}.

\textbf{Grad-CAM} is a way of visualizing the model decision. With necessity of model explanation, several works have developed method of visualizing saliency-maps via using gradient of model decisions~\cite{saliency, CAM}, and one of the state-of-the-art approaches \cite{ECT} leveraged the idea of explanation as the surrogate of causality by exploiting gradient-based saliency-map explanation consistency training. On top of this feasibility test on using saliency-map consistency training, we here show better application by using more advanced method, for example, Grad-CAM.

\section{Method}
The main idea of our method is using Grad-CAM for consistency training. As the second and fourth columns in \autoref{fig:overview} show, the model outputs different Grad-CAM results from an image with perturbation. Following smoothness assumption, we train the model to output similar results not only on labels but also on Grad-CAM result with specified target layer.

The objective of conventional SSL training can be written as: given model $ f( \cdot ) $ with labeled dataset $ \{X_L, Y_L\} = \{(\mathbf{x}_i, y_i)\}_{i=1}^N $ and unlabeled dataset $ X_U = \{ \mathbf{x}_{j} \}_{j=1}^M $,

\begin{align*}
    & \underset{\theta \in \Theta }
        {\operatorname{min}} \frac{1}{N}  \sum_{i=1}^{N} l( f(\mathbf{x}_i;\theta) ,y_i ) 
                    + \alpha \frac{1}{U}  \sum_{j=1}^{M} k( \mathbf{x}_j , Aug(\mathbf{x}_j) ) \\
\end{align*} where $ l, k $ denotes empirical supervised loss and consistency loss (e.g., $ L2 $-loss), respectively, and $ Aug(\cdot) $ is a perturbation function (e.g., adversarial augmentation \cite{VAT}). 

Here, we adopt consistency training with explanation, and let the objective function $ k( \cdot ) $ be $ \frac{1}{M} \sum_{j=1}^M ( E(\mathbf{x}_j;\theta, L) - E( Aug(\mathbf{x}_j);\theta, L) )^2 $ where $ E (\mathbf{x};\theta, L) $ captures the Grad-CAM result from the input data \(\mathbf{x}\) with model parameter $\theta$ and the target layer $L$. 

To be more specific, we add additional label consistency loss, meaning negative of cross entropy between predicted labels by the model from the original data and augmented data \( \sum_{j=1}^M f( \mathbf{x}_j ) log( f( Aug( \mathbf{x}_j ) ) )  \). More detailed algorithm is below.

\begin{algorithm}
\caption{Consistency training with Grad-CAM loss}\label{alg:cap}
\begin{algorithmic}
\State Initialize model function $ f(\cdot) $ with parameter $ \theta $
\State Set training coefficients $\alpha, \beta, \gamma$ 
\State Set target layer $ L $ for Grad-CAM result
\For {epoch 1, E}
\For {sample 1, N}
\State Sample labeled data $ (\mathbf{x}_l, y_l) $ from $ \{ X_L, Y_L \} $ 
\State Compute supervised loss $ L_{S} $ from $ f(\mathbf{x}_l) $ and $ y_l $
\EndFor
\For {sample 1, N}          \Comment{same number of samples}
\State Sample unlabeled data $ \mathbf{x}_u $ from iterator of  $ \{ X_U \} $
\State Set augmented data $\mathbf{z}_u = Aug(\mathbf{x}_u) $
\State Compute pseudo-label loss $ L_{P} $ from pseudo-label $ f(\mathbf{x}_u) $ and prediction $ f(\mathbf{z}_u)$
\State Compute Grad-CAM loss $ L_{G} $ from $ E(\mathbf{x}_u;\theta, L) $ and $ E( Aug(\mathbf{z}_u);\theta, L) $ 
\State Set overall loss $L = \alpha L_{S} + \beta L_{P} + \gamma L_{G} $
\State Update parameter $ \theta  \gets  \theta - \nabla L $
\EndFor
\EndFor
\end{algorithmic}
\end{algorithm}

As the former layers and the latter layers learn features with different scales, i.e., respectively local and global regions, we attempted the model to be trained to output similar output in different scales by targeting different layers of the baseline model.

\begin{figure}[!tb]
 	\centering
    	\includegraphics[width=8cm]{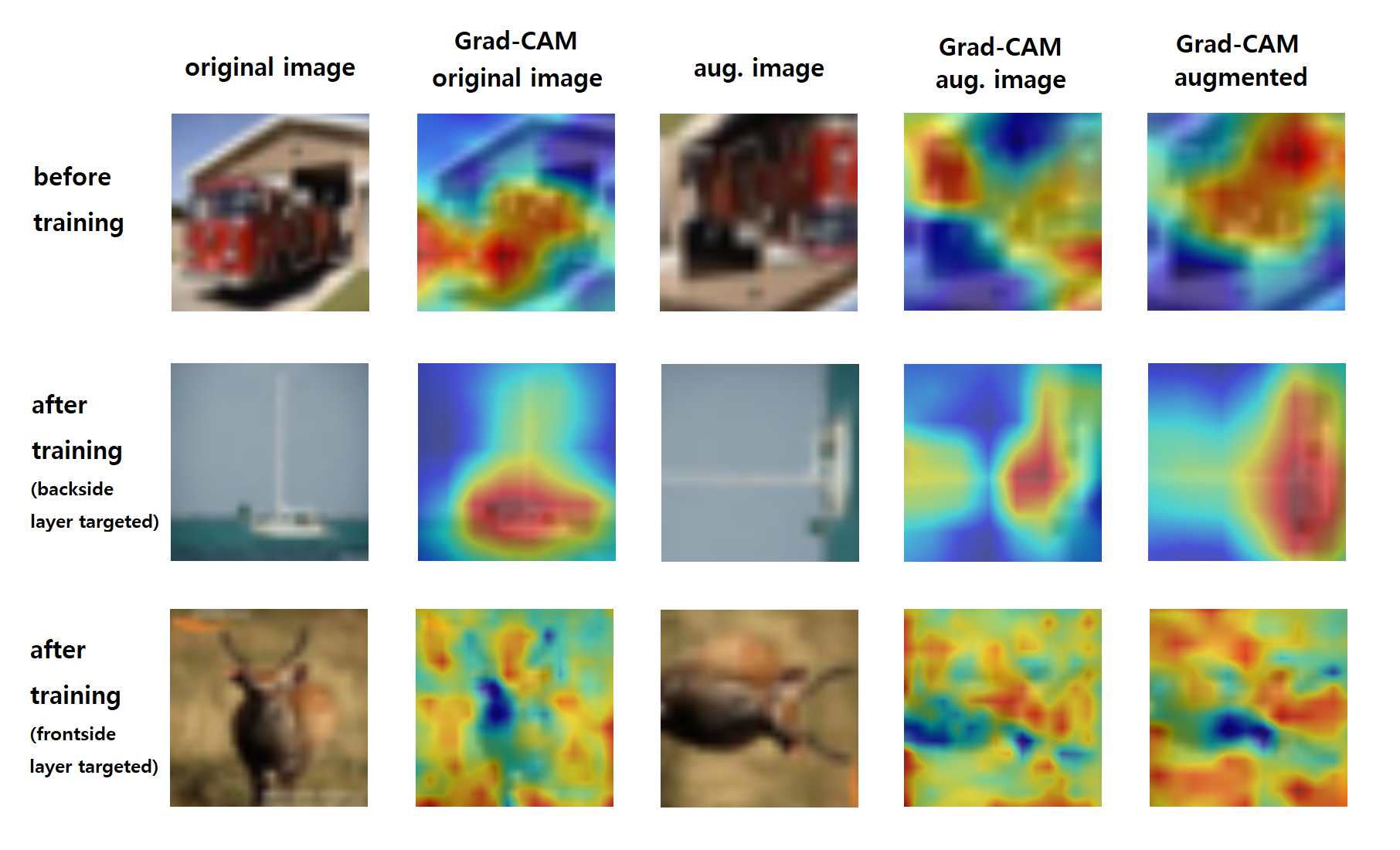}
    	\caption{ How Grad-CAM result changes after training. The first row shows the differences between Grad-CAM result from original image and augmented image. The second and third row show how the results become similar. The last column shows the augmented version of Grad-CAM result from original image, so this was used as the pseudo-label. The difference between the second and the third row is the target layer. Here, the crop and rotation are only used for augmentation. }
      	\label{fig:training}
 \end{figure}

\section{Result}

\subsection{Experiment setting}
Here, we will explain the experiment setting details, especially on the dataset. ResNet50 was used for the baseline model during the experimentation. The model is trained with CIFAR-10 dataset \cite{cifar10} for image classification.

We used 1/5 of the training data for the validation dataset, and divided the training dataset into labeled training dataset and unlabeled training dataset. To test different ratios, the number of labeled training data was fixed to 5000, so each label contains 100 images on average. For the target ratios of unlabeled to labeled, 0.5, 1.0, 2.0 were selected, meaning 2500, 5000, 10000 images, respectively.

For image augmentation, we used crop, rotation, and color-jitters. In terms of fair comparison, we trained the baseline model with randomly augmented labeled data.

\begin{figure}[b]
 	\centering
    	\includegraphics[width=8cm]{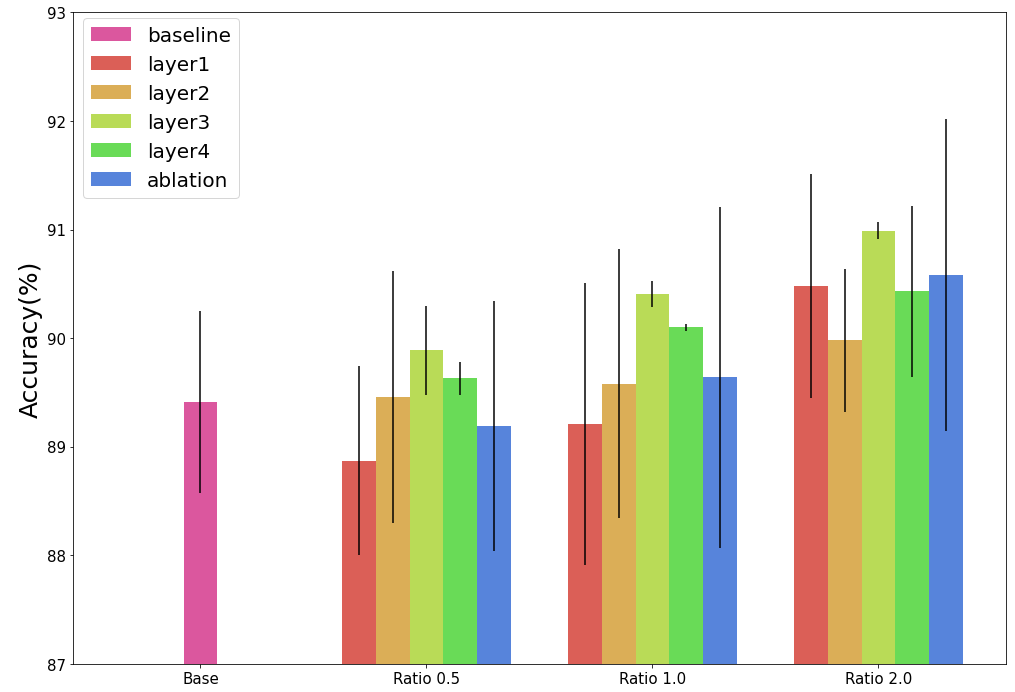}
    	\caption{ Results from different ratios and target layers with our method of using consistent training with Grad-CAM loss on image classification task. The average results and standard deviations from several tests are stated. Results from ablation study are also shown together for comparison. }
      	\label{fig:result}
 \end{figure}

\subsection{Performance test}

\autoref{fig:training}  shows, the Grad-CAM result from two different images became similar as the model gets trained. Comparing to the baseline model, our method showed improved accuracy rate result on image classification task. \autoref{table:training} states the empirical statistics of Grad-CAM consistency loss with test on validation set. 

With more unlabeled data, model showed better improvements. This indicates that the use of more training data results in better generalization. 

The target layer stated in the \autoref{table:training} indicates the last layer of the block (following the implementation of pytorch library \cite{pytorch_resnet}). The model showed best performance with the target layer 3, and showed a trend of more improvements as the target goes to the backside. We argue that this result is not arbitrary, but has strong correlation with the label distribution of the dataset.  

\begin{table}[t]
\begin{center}
\caption{Performance test on consistency training with Grad-CAM}
\begin{tabular}{|c||c|c|c|c|c|}
\hline
\multirow{2}{*}{\textbf{Ratio}}
                         & \multicolumn{5}{c|}{\textbf{Accuracy [\%]}} \\ \cline{2-6} 
                         & \textbf{Layer1} & \textbf{Layer2}  & \textbf{Layer3} & \textbf{Layer4} & \textbf{Baseline} \\ 
\hline
\multirow{2}{1em} {0.5}  & 88.87           & 89.46            & 89.89           & 89.63           &                   \\ 
                         & $\pm$ 0.88      & $\pm$ 1.58       & $\pm$ 0.41      & $\pm$ 0.15      &                   \\
\cline{1-5}
\multirow{2}{1em} {1.0}  & 89.33           & 89.59            & 90.41           & 90.10           &  89.61            \\
                         & $\pm$ 1.47      & $\pm$ 1.24       & $\pm$ 0.12      & $\pm$ 0.03      &  $\pm$ 0.84       \\
\cline{1-5}
\multirow{2}{1em} {2.0}  & 90.48           & 89.98            & \textbf{90.99}  & 90.43           &                   \\
                         & $\pm$ 0.88      & $\pm$ 0.88       & $\pm$ 0.88      & $\pm$ 0.88      &                   \\
\hline  
\end{tabular}
\label{table:training}
\end{center}
\end{table}

\subsection{Ablation study}

\autoref{table:ablation} shows the result of the ablation study, examining how much Grad-CAM consistency along affected the training procedure. For the baseline in this ablation study, we used the combination of supervision loss $ L_{S} $ and pseudo-label loss $ L_{P} $.

To note on the baseline, as we used the loss function of cross-entropy loss, the result was worse than using only supervision loss. Nevertheless, the comparison states the exploitation of Grad-CAM loss guides the model appropriately. We are planning to apply different loss function for the pseudo-label loss. 

The best results in the second column of the table shows the best results by using all three losses containing Grad-CAM loss $ L_{G} $ among all different ratios and target layers, showing that the use of Grad-CAM loss results in the improvement.

\section{Conclusion}

We have showed that the improvement can be achieved by using Grad-CAM consistency loss comparing to the baseline model or consistency training with only using pseudo-label. With trends in the result shows, the model showed better improvements with more unlabeled data used.

Unlike the latest studies \cite{ECT} \cite{cam_ct} , we could develop better use-case of the training method with respect to the adjustability. Here, we address some discussions on how the model could get generalized and how our method can be applied for training different data with enhanced adjustability.

\subsection{On generalization}
With little examination, computing Grad-CAM can be viewed as a weakly-supervised semantic segmentation. Thus, the consistency loss with Grad-CAM result may work as slightly different version of the loss used in \cite{consistent-semantic}; this work presents a method of consistency training for semantic segmentation task. Thus, Grad-CAM producing task would have worked as an auxiliary task, resulting better generalization on the image classification task also.

\subsection{On adjustability}
Our hypothesis on how targeting different layer matters is: as the former layers learn features from local regions, while the latter layers learn features from larger regions, training model to produces similar attention on the latter layer would affect the model to produce similar results in global scale. In other words, we believe when the label classes are hugely different (i.e., cat vs. dog) targeting latter layers are efficient, while when the label classes are locally different (i.e., leopard vs. cheetah) targeting front layers will be appropriate. 

Although this hypothesis remains to be ensured with more experiments, we suggests our method to be used in variety, within different experimentation such as on different dataset having different label distributions. 

\begin{table}[t]
\begin{center}
\caption{Grad-CAM consistency training ablation test}
\begin{tabular}{ |c|c|c|c| } 
\hline
\multirow{2}{*}{\textbf{Ratio}} & \multicolumn{2}{c|}{\textbf{Accuracy [\%]}} \\\cline{2-3} 
& \textbf{Best} & \textbf{Ablation}  \\ 
\hline
0.5 & 89.89 & 89.19 $\pm$ 1.16 \\ 
\hline
1.0 & 90.41 & 89.64 $\pm$ 1.57 \\ 
\hline
2.0 & 90.99 & 90.56 $\pm$ 1.44 \\ 
\hline
\end{tabular}
\end{center}
\label{table:ablation}
\end{table}

\bibliographystyle{ms}
\bibliography{ms} 

\begin{thebibliography}{10}
\providecommand{\url}[1]{#1}
\csname url@samestyle\endcsname
\providecommand{\newblock}{\relax}
\providecommand{\bibinfo}[2]{#2}
\providecommand{\BIBentrySTDinterwordspacing}{\spaceskip=0pt\relax}
\providecommand{\BIBentryALTinterwordstretchfactor}{4}
\providecommand{\BIBentryALTinterwordspacing}{\spaceskip=\fontdimen2\font plus
\BIBentryALTinterwordstretchfactor\fontdimen3\font minus
  \fontdimen4\font\relax}
\providecommand{\BIBforeignlanguage}[2]{{%
\expandafter\ifx\csname l@#1\endcsname\relax
\typeout{** WARNING: IEEEtran.bst: No hyphenation pattern has been}%
\typeout{** loaded for the language `#1'. Using the pattern for}%
\typeout{** the default language instead.}%
\else
\language=\csname l@#1\endcsname
\fi
#2}}
\providecommand{\BIBdecl}{\relax}
\BIBdecl

\bibitem{dl}
Y.~LeCun, Y.~Bengio, and G.~Hinton, ``Deep learning,'' \emph{Nature}, vol. 521,
  no. 7553, pp. 436--444, 2015.

\bibitem{SSL}
J.~Ratsaby and S.~S. Venkatesh, ``Learning from a mixture of labeled and
  unlabeled examples with parametric side information,'' in \emph{Proceedings
  of the Eighth Annual Conference on Computational Learning Theory}, ser. COLT
  '95.\hskip 1em plus 0.5em minus 0.4em\relax New York, NY, USA: Association
  for Computing Machinery, 1995, p. 412–417.

\bibitem{consistency}
M.~Sajjadi, M.~Javanmardi, and T.~Tasdizen, ``Regularization with stochastic
  transformations and perturbations for deep semi-supervised learning,'' in
  \emph{Proceedings of the 30th International Conference on Neural Information
  Processing Systems}, ser. NIPS'16.\hskip 1em plus 0.5em minus 0.4em\relax Red
  Hook, NY, USA: Curran Associates Inc., 2016, p. 1171–1179.

\bibitem{ECT}
T.~Han, W.-W. Tu, and Y.-F. Li, ``Explanation consistency training:
  Facilitating consistency-based semi-supervised learning with
  interpretability,'' in \emph{Proceedings of the AAAI Conference on Artificial
  Intelligence}, vol.~35, no.~9, 2021, pp. 7639--7646.

\bibitem{cam_ct}
H.~Guo, K.~Zheng, X.~Fan, H.~Yu, and S.~Wang, ``Visual attention consistency
  under image transforms for multi-label image classification,'' in
  \emph{Proceedings of the IEEE/CVF Conference on Computer Vision and Pattern
  Recognition}, 2019, pp. 729--739.

\bibitem{gradcam}
R.~R. Selvaraju, M.~Cogswell, A.~Das, R.~Vedantam, D.~Parikh, and D.~Batra,
  ``Grad-cam: Visual explanations from deep networks via gradient-based
  localization,'' in \emph{Proceedings of the IEEE international conference on
  computer vision}, 2017, pp. 618--626.

\bibitem{resnet}
K.~He, X.~Zhang, S.~Ren, and J.~Sun, ``Deep residual learning for image
  recognition,'' in \emph{Proceedings of the IEEE conference on computer vision
  and pattern recognition}, 2016, pp. 770--778.

\bibitem{SSL_smooth}
A.~Goldberg, X.~Zhu, A.~Singh, Z.~Xu, and R.~Nowak, ``Multi-manifold
  semi-supervised learning,'' in \emph{Proceedings of the Twelth International
  Conference on Artificial Intelligence and Statistics}, ser. Proceedings of
  Machine Learning Research, D.~van Dyk and M.~Welling, Eds., vol.~5.\hskip 1em
  plus 0.5em minus 0.4em\relax Hilton Clearwater Beach Resort, Clearwater
  Beach, Florida USA: PMLR, 16--18 Apr 2009, pp. 169--176.

\bibitem{SSL_entropy}
Y.~Grandvalet and Y.~Bengio, ``Semi-supervised learning by entropy
  minimization,'' in \emph{Advances in Neural Information Processing Systems},
  L.~Saul, Y.~Weiss, and L.~Bottou, Eds., vol.~17.\hskip 1em plus 0.5em minus
  0.4em\relax MIT Press, 2005.

\bibitem{SSL_manifold}
M.~Belkin and P.~Niyogi, ``Semi-supervised learning on riemannian manifolds,''
  in \emph{Machine Learning}, 2004, pp. 209--239.

\bibitem{MixMatch}
D.~Berthelot, N.~Carlini, I.~J. Goodfellow, N.~Papernot, A.~Oliver, and
  C.~Raffel, ``Mixmatch: {A} holistic approach to semi-supervised learning,''
  \emph{CoRR}, vol. abs/1905.02249, 2019.

\bibitem{meanteacher}
A.~Tarvainen and H.~Valpola, ``Mean teachers are better role models:
  Weight-averaged consistency targets improve semi-supervised deep learning
  results,'' \emph{arXiv preprint arXiv:1703.01780}, 2017.

\bibitem{VAT}
T.~Miyato, S.-i. Maeda, M.~Koyama, and S.~Ishii, ``Virtual adversarial
  training: a regularization method for supervised and semi-supervised
  learning,'' \emph{IEEE transactions on pattern analysis and machine
  intelligence}, vol.~41, no.~8, pp. 1979--1993, 2018.

\bibitem{fixmatch}
A.~Kurakin, C.-L. Li, C.~Raffel, D.~Berthelot, E.~D. Cubuk, H.~Zhang, K.~Sohn,
  N.~Carlini, and Z.~Zhang, ``Fixmatch: Simplifying semi-supervised learning
  with consistency and confidence,'' in \emph{NeurIPS}, 2020.

\bibitem{saliency}
K.~Simonyan, A.~Vedaldi, and A.~Zisserman, ``Deep inside convolutional
  networks: Visualising image classification models and saliency maps,''
  \emph{arXiv preprint arXiv:1312.6034}, 2013.

\bibitem{CAM}
B.~Zhou, A.~Khosla, A.~Lapedriza, A.~Oliva, and A.~Torralba, ``Learning deep
  features for discriminative localization,'' in \emph{Proceedings of the IEEE
  conference on computer vision and pattern recognition}, 2016, pp. 2921--2929.

\bibitem{cifar10}
A.~Krizhevsky, V.~Nair, and G.~Hinton, ``Cifar-10 (canadian institute for
  advanced research).''

\bibitem{pytorch_resnet}
A.~Paszke, S.~Gross, F.~Massa, A.~Lerer, J.~Bradbury, G.~Chanan, T.~Killeen,
  Z.~Lin, N.~Gimelshein, L.~Antiga, A.~Desmaison, A.~Kopf, E.~Yang, Z.~DeVito,
  M.~Raison, A.~Tejani, S.~Chilamkurthy, B.~Steiner, L.~Fang, J.~Bai, and
  S.~Chintala, ``Pytorch: An imperative style, high-performance deep learning
  library,'' in \emph{Advances in Neural Information Processing Systems 32},
  H.~Wallach, H.~Larochelle, A.~Beygelzimer, F.~d\textquotesingle
  Alch\'{e}-Buc, E.~Fox, and R.~Garnett, Eds.\hskip 1em plus 0.5em minus
  0.4em\relax Curran Associates, Inc., 2019, pp. 8024--8035.

\bibitem{consistent-semantic}
Y.~Ouali, C.~Hudelot, and M.~Tami, ``Semi-supervised semantic segmentation with
  cross-consistency training,'' in \emph{Proceedings of the IEEE/CVF Conference
  on Computer Vision and Pattern Recognition}, 2020, pp. 12\,674--12\,684.

\end{thebibliography}

\end{document}